\documentclass[preprint,3p,times]{elsarticle}

\usepackage{amssymb}
\usepackage{amsmath}

\usepackage{hyperref}
\usepackage{booktabs}
\usepackage{multirow}
\usepackage{wrapfig}
\usepackage{subfig}
\usepackage{url}
\usepackage{xcolor}
\usepackage[table]{xcolor}
\usepackage{algpseudocode}
\usepackage{algorithm}
\DeclareMathOperator*{\argmin}{arg\,min}
\usepackage{amssymb}

\usepackage{pifont}
\begin{document}

\begin{frontmatter}



\title{Resource-efficient Layer-wise
Federated Self-supervised Learning}




\author[label1]{Ye Lin Tun}
\ead{yelintun@khu.ac.kr}
\author[label1]{Chu Myaet Thwal}
\ead{chumyaet@khu.ac.kr}
\author[label1]{Huy Q. Le}
\ead{quanghuy69@khu.ac.kr}
\author[label2]{Minh N. H. Nguyen}
\ead{nhnminh@vku.udn.vn}
\author[label1]{Eui-Nam Huh}
\ead{johnhuh@khu.ac.kr}
\author[label1]{Choong Seon Hong\corref{cor1}}
\ead{cshong@khu.ac.kr}

\affiliation[label1]{organization={Department of Computer Science and Engineering, Kyung Hee University},
            city={Yongin-si},
            state={Gyeonggi-do 17104},
            country={South Korea}}
            
\affiliation[label2]{organization={Vietnam - Korea University of Information and Communication Technology},
            city={Danang},
            country={Vietnam}} 

\cortext[cor1]{Corresponding author}

\begin{abstract}

\vspace{2em}
\noindent
Many studies integrate federated learning (FL) with self-supervised learning (SSL) to take advantage of raw data distributed across edge devices.
However, edge devices often struggle with high computational and communication costs imposed by SSL and FL algorithms. 
With the deployment of more complex and large-scale models, these challenges are exacerbated.
To tackle this, we propose Layer-Wise Federated Self-Supervised Learning (\mbox{LW-FedSSL}), which allows edge devices to incrementally train a small part of the model at a time.
Specifically, in LW-FedSSL, training is decomposed into multiple stages, with each stage responsible for only a specific layer of the model.
Since only a portion of the model is active for training at any given time, LW-FedSSL significantly reduces computational requirements.
Additionally, only the active model portion needs to be exchanged between the FL server and clients, reducing communication overhead.
This enables LW-FedSSL to jointly address both computational and communication challenges of FL client devices.
It can achieve up to a $3.34 \times$ reduction in memory usage, $4.20 \times$ fewer computational operations (giga floating point operations, GFLOPs), and a $5.07 \times$ lower communication cost while maintaining performance comparable to its end-to-end training counterpart. 
Furthermore, we explore a progressive training strategy called Progressive Federated Self-Supervised Learning (\mbox{Prog-FedSSL}), which offers a $1.84\times$ reduction in GFLOPs and a $1.67\times$ reduction in communication costs while maintaining the same memory requirements as end-to-end training.
Although the resource efficiency of \mbox{Prog-FedSSL} is lower than that of \mbox{LW-FedSSL}, its performance improvements make it a viable candidate for FL environments with more lenient resource constraints.
\vspace{2em}
\end{abstract}


\end{frontmatter}



\vspace{2em}

\section{Introduction}

\vspace{1em}

A significant portion of real-world data valuable for practical AI applications is distributed across edge devices, often subject to privacy constraints.
Decentralized learning approaches like federated learning (FL) \citep{konevcny2016federated,mcmahan2017communication} address these challenges by enabling privacy-preserving collaborative model training.
FL systems operate by collecting trained model parameters from edge devices (a.k.a., clients) instead of their data, thereby preserving data privacy.
This approach is particularly important for privacy-sensitive data such as medical records or financial information.
However, much of the data on edge devices remains unlabeled, limiting the effectiveness of traditional supervised learning methods. 
To tackle this, self-supervised learning (SSL) strategies \citep{chen2020simple,he2020momentum,chen2021exploring} allow models to learn from raw data by generating their own supervisory signals without requiring labels.
Typically, SSL methods rely on centralized training schemes, which often involve exhaustive data collection to build a centrally stored dataset. 
Privacy concerns and the distributed nature of edge data make this centralized approach infeasible in many real-world scenarios.
Consequently, federated self-supervised learning (FedSSL) has recently emerged as a promising approach that combines the strengths of FL and SSL, enabling models to leverage unlabeled data distributed across clients while preserving privacy  \citep{zhuang2021collaborative,li2023mocosfl,wang2023does}.

Self-supervised learning (SSL), particularly with state-of-the-art models like Transformers \citep{vaswani2017attention,dosovitskiy2021an}, imposes substantial computational demands. 
Unfortunately, clients in an FL environment often operate with limited computational and communication resources. 
Many clients may lack the necessary resources to participate in a conventional end-to-end FedSSL process. 
As a result, data from these clients is excluded from the training process, potentially degrading overall model performance.
Therefore, it is crucial to explore resource-efficient FedSSL approaches that enable every FL client to participate in collaborative training.
A straightforward way to reduce computational demands on low-memory devices is to conduct the training with a smaller batch size. 
However, using a small batch size in SSL can diminish the quality of learned representations and compromise performance \citep{chen2020simple,li2023mocosfl}. 
Moreover, simply reducing the batch size does not lower the FL communication costs, as clients still need to exchange the full model with the server.
On the other hand, using a large batch size would exclude low-memory devices from participating in the FL process.

To address these challenges, we introduce LW-FedSSL, a layer-wise training approach where a model is systematically trained one layer at a time.
The term ``\textit{layer}'' in this context can refer to either an individual layer or a block of multiple layers within the model.
In LW-FedSSL, the training process is divided into multiple stages, with each stage focusing on a specific portion of the model.
At the beginning of each stage, a new layer is sequentially added to the model, while all previously trained layers from prior stages (if any exists) are frozen, preventing further updates.
Only the newly added layer remains active for training, significantly reducing computational requirements for FL clients.
Additionally, since only the active layer within a stage needs to be exchanged between the server and FL clients, communication bottlenecks are substantially reduced.
As training progresses with each stage, the model depth increases gradually, enabling an incremental and efficient learning process.
We also explore a progressive training strategy, Prog-FedSSL, which follows a similar approach to layer-wise training.
Like LW-FedSSL, Prog-FedSSL divides the training process into multiple stages, with each stage sequentially adding a new layer to the model.
However, instead of freezing previously trained layers, Prog-FedSSL keeps all existing layers active during each stage, allowing the entire sub-model to continue updating.
As a result, Prog-FedSSL may require more computational and communication resources than LW-FedSSL, but it has the potential to improve model performance.



Our contributions can be summarized as follows:
\begin{itemize}
    
    \item We introduce \mbox{LW-FedSSL}, a novel layer-wise federated self-supervised learning framework designed to significantly enhance the resource efficiency of FL clients.
    \mbox{LW-FedSSL} can compete with conventional end-to-end training counterparts while simultaneously reducing both computational and communication costs.
    
    \item We explore \mbox{Prog-FedSSL}, a progressive training strategy for federated self-supervised learning, that holds the potential to enhance performance.
    Given its lower resource efficiency compared to \mbox{LW-FedSSL}, \mbox{Prog-FedSSL} is better suited for FL environments with more flexible resource constraints.
    
    \item  We provide an in-depth exploration of LW-FedSSL and Prog-FedSSL within the FL paradigm, considering various empirical aspects such as computational efficiency, communication overhead, and model performance.
    We present a thorough evaluation of our proposed approaches across different FL settings and benchmarks.

\end{itemize}

\section{Background and Related Work}

In this section, we provide a brief overview of key concepts in federated learning and self-supervised learning, along with a discussion of related work.

\subsection{Federated Learning}

\begin{figure}[ht]
    \centering
    \includegraphics[width=0.4\linewidth]{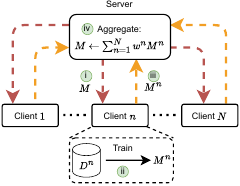}
    \caption{Four key steps in a single FL communication round. (i) The server distributes a base global model $M$ to all participating clients. (ii) Each client $n \in [1,N]$  trains the model on its local dataset $D^n$ to produce a local model $M^n$. 
    (iii) The local models are then transmitted back to the server. 
    (iv) Finally, the server aggregates the received local models using weighted averaging to update the global model: $M \leftarrow \sum^N_{n=1} w^n M^n$ \cite{mcmahan2017communication}.}
    \label{fig:fl}
\end{figure}

A typical FL process involves a central coordinating server and a set of client devices, each storing private data \citep{mcmahan2017communication}. 
The goal of FL is to train a global model $M$, by learning from the data residing on clients while ensuring their privacy. 
The FL objective can be expressed as:
\begin{equation}
    M^* = \argmin_{M} \sum_{n=1}^N w^n \mathcal{L} (M, D^n),
    \label{eqn:fl_obj}
\end{equation}
where $M^*$  is the optimal global model after training, $N$ is the number of clients, and $\mathcal{L}$ represents the local loss function.
Here, $w^n=|D^n|/|D|$ is the weight assigned to the $n$-th client, with $|D^n|$ denoting the number of samples in the local dataset $D^n$, and $|D|$ denoting the total number of samples across all clients, i.e., ${D=\bigcup_{n=1}^{N} D^n}$.
As shown in Fig.~\ref{fig:fl}, a single FL communication round consists of four main steps, which are repeated for a total of $R$ rounds.

\subsection{Self-supervised Learning}

\begin{figure}[ht]
    \centering
    \includegraphics[width=0.4\linewidth]{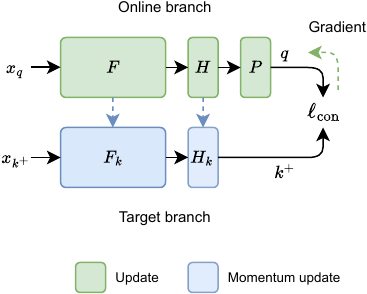}
    \caption{Self-supervised learning with MoCoV3 \citep{9711302}. We omit the negative samples for clarity.}
    \label{fig:ssl}
\end{figure}

Self-supervised learning (SSL) offers a way to utilize unlabeled data for model training, which is often more readily available than labeled data. 
SSL generates its own supervisory signals to learn valuable representations from the unlabeled data.
In this work, we primarily use MoCoV3~\citep{9711302} as the SSL approach.\footnote{
Unlike in \citep{9711302}, we train the patch projection layer.}
As shown in Fig.~\ref{fig:ssl}, MoCoV3 features a Siamese network structure, consisting of an actively trained online branch, and a target branch. 
The online branch includes an encoder $F$, a projection head $H$, and a prediction head $P$. 
Meanwhile, the target branch is a moving-averaged copy of the online branch, consisting of a momentum encoder $F_k$ and a momentum projection head $H_k$. 

Given an input sample $x$, it undergoes augmentation, creating two views, $x_q$ and $x_{k^+}$. These views are then fed into the online and target branches, respectively, to obtain representations $q$ and its corresponding positive pair, $k^+$. Likewise, negative pairs, $k^-$, are derived from other samples within the same batch through the target branch. The online branch is trained using the \mbox{InfoNCE} loss \citep{van2018representation}, defined as:

\begin{equation}
    \label{eqn:infonce}
    \ell_\text{con}(q, k, \tau) = - \log \frac{\exp (q \cdot k^+/ \tau)}{\exp(q \cdot k^+/ \tau) + \sum_{j=1}^K \exp(q \cdot k_j^- / \tau)},
\end{equation}
where $k = (k^+, \{k_j^-\}_{j=1}^K)$ and $\tau$ is the temperature parameter. Using a momentum parameter $\mu$, the target branch is updated as: $F_k \leftarrow \mu F_k + (1-\mu)F$ and $H_k \leftarrow \mu H_k + (1-\mu)H$. 

\subsection{Federated Self-supervised Learning}

While most studies on FL focus on supervised learning \citep{mcmahan2017communication, li2020federated, Li_2021_CVPR}, such approaches place the burden of maintaining labeled data on each device owner, which can be challenging in many real-world settings.
Consequently, client devices often lack labeled data due to limitations in data collection, labeling resource constraints, and insufficient annotation expertise. 
Federated self-supervised learning (FedSSL) extends the applicability of FL to new domains where labeled data is scarce or difficult to acquire \citep{zhuang2022divergence, saeed2020federated, ek2022federated}. 
Many FedSSL studies build upon state-of-the-art SSL frameworks, such as SimCLR~\citep{chen2020simple}, MoCo~\citep{he2020momentum}, and BYOL~\citep{grill2020bootstrap}.
These studies adapt SSL techniques to address specific FL tasks \citep{wu2022distributed, dong2021federated} or tackle FL challenges at hand \citep{makhija2022federated, wu2021decentralized, zhang2023federated, zhuang2021collaborative}.
In the context of FedSSL, the local dataset $D^n$ within a client $n$ is unlabeled, and thus, the local training process relies on a form of SSL loss to train the local model~$M^n$. 
When using MoCoV3 as the SSL framework, the local model consists of an encoder $F^n$ and its accompanying MLP heads, $H^n$ and $P^n$.

\subsection{Layer-wise Training}
\label{sec:related_work}

Layer-wise training was initially introduced in the context of deep belief networks and restricted Boltzmann machines \citep{10.1162/neco.2006.18.7.1527,10.5555/2976456.2976476}. 
The original motivation behind its development was to address challenges associated with training deep neural networks, such as vanishing gradients. 
While its usage has become less common over the years, we believe it holds great potential to be rediscovered as a highly resource-efficient strategy within FL systems.
Incrementally training a model one layer at a time offers a significant resource-saving advantage over end-to-end training.
Despite its potential benefits, layer-wise training remains largely unexplored in FL, with very few related studies \citep{wang2022progfed, pengfei2023towards}.

The study in \cite{huo2021incremental} investigates layer-wise SSL in the speech domain. 
Although it was intended for an on-device training scenario, the authors mainly examined it in a centralized setting.
A short study \cite{pengfei2023towards} on federated layer-wise learning (FLL) introduces a depth dropout technique that randomly drops frozen layers during training to reduce resource overhead.
However, FLL is only examined with a large number of communication rounds in each training stage (ranging from 4k to 12k), potentially placing significant strain on clients.
Additionally, the exploration of FLL is limited in scope, missing comprehensive evaluations across diverse FL settings and benchmark datasets.
In contrast to these studies, our work aims to offer a more thorough and in-depth exploration of layer-wise training through LW-FedSSL.
One study \cite{wang2022progfed} investigates the progressive training approach for FL known as ProgFed. 
Although similar to layer-wise training in gradually increasing the model depth, ProgFed~\cite{wang2022progfed} differs by training all existing layers at each stage. 
Since ProgFed primarily focuses on supervised learning tasks that require labeled data, the nature of progressive training within the FedSSL paradigm remains unexplored.
To fill this gap, we introduce a progressive training strategy for FedSSL, referred to as \mbox{Prog-FedSSL}.

\begin{figure*}[ht]
    \centering
    \includegraphics[width=.98\linewidth]{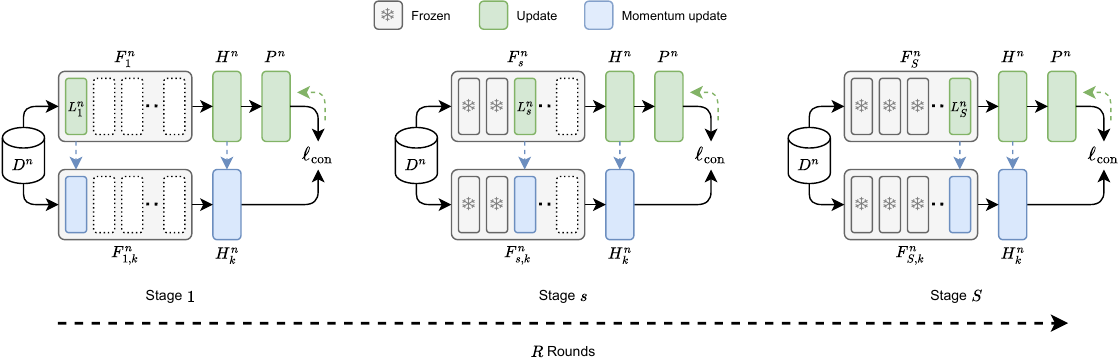}
    \caption{LW-FedSSL: Local training process across different stages for the $n$-th client. At the beginning of each stage $s \in [1,S]$, a new layer $L_s$ is sequentially added to the previous encoder $F_{(s-1)}$, increasing its depth. During stage $s$, only the corresponding layer $L_s$ is actively updated, while all prior layers (i.e., $L_1$ to $L_{(s-1)}$) are kept frozen.}
    \label{fig:lwssl}
\end{figure*}


\begin{algorithm}[ht]
\caption{LW-FedSSL (\emph{Server-side})}
\label{alg:lwfedssl_server}
\textbf{Input:} encoder $F_0$, projection head $H$, prediction head $P$, number of stages $S$, number of rounds per stage $R_s$ where $s \in [1,S]$ \\
\textbf{Output:} encoder $F_S$

\begin{algorithmic}[1]
    \State \textbf{Server executes:}
    \State Distribute $F_0$ to clients
    \For{stage $s=1,2,\dots, S$}
        \State Initialize new layer: $L_s$
        \For{round $r=1,2,\dots, R_s$}
            \For{client $n=1,2,\dots,N$ in parallel}
                \vspace{0.2em}
                \State $L_s^n, H^n, P^n \leftarrow \text{Train}(n, L_s, H, P)$
                \vspace{0.2em}
                \State $w^n = \frac{|D^n|}{|D|}$, where $D=\bigcup_{n=1}^{N} D^n$
            \EndFor
            \vspace{0.2em}
            \State $L_s \leftarrow \sum_{n=1}^N w^n L_s^n$
            \vspace{0.2em}
            \State $H \leftarrow \sum_{n=1}^N w^n H^n$
            \vspace{0.2em}
            \State $P \leftarrow \sum_{n=1}^N w^n P^n$
        \EndFor
        \State Distribute $L_s$ to clients
        \State $F_s \leftarrow$ Attach $L_s$ to $F_{(s-1)}$
    \EndFor
    \State return $F_S$
\end{algorithmic}
\end{algorithm}

\begin{algorithm}[ht]
\caption{LW-FedSSL (\emph{Client-side})}
\label{alg:lwfedssl_client}
\textbf{Input:} local dataset $D^n$, encoder $F^n_{(s-1)}$, number of local epochs $E$, momentum $\mu$, temperature $\tau$\\
\textbf{Output:} layer $L^n_s$, projection head $H^n$, prediction head $P^n$ 
\begin{algorithmic}[1]
    \State \textbf{Client executes:} Train$(n, L_s, H, P)$:
        \vspace{0.2em}
        \State Initialize: $L^n_s \leftarrow L_s$, $H^n \leftarrow H$, $P^n \leftarrow P$
        \vspace{0.2em}
        \State $F^n_s \leftarrow$ Attach trainable $L^n_s$ to frozen $F^n_{(s-1)}$
        \vspace{0.2em}
        \State Target branch: $F^n_{s,k} \leftarrow F^n_s$, $H^n_k \leftarrow H^n$ 
        \vspace{0.2em}
        \For{epoch $e=1,2,\dots,E$}
            \For{each batch $x \in D^n$}
                \State $x_1 \leftarrow \text{{Augment}}(x)$
                \vspace{0.2em}
                \State $x_2 \leftarrow \text{{Augment}}(x)$
                \vspace{0.2em}
                
                \State $q_1 \leftarrow P^n(H^n(F^n_s(x_1)))$
                \vspace{0.2em}
                \State $q_2 \leftarrow P^n(H^n(F^n_s(x_2)))$
                \vspace{0.2em}

                \State $k_1 \leftarrow H^n_k(F^n_{s,k}(x_1))$
                \vspace{0.2em}
                \State $k_2 \leftarrow H^n_k(F^n_{s,k}(x_2))$
                \vspace{0.2em}




                \State $\mathcal{L} \leftarrow \ell_\text{con}(q_1, k_2, \tau) + \ell_\text{con}(q_2, k_1, \tau)$
                \Comment{Local loss}
                \vspace{0.2em}

                \State $F^n_s, H^n, P^n \leftarrow$ Update with $\nabla\mathcal{L}$
                \vspace{0.2em}

                \State $F^n_{s,k}, H^n_k,\leftarrow$ Momentum update with $\mu, F^n_s, H^n$
                \vspace{0.2em}
            
            \EndFor
        \EndFor
        \State $L^n_s \leftarrow \text{Get active layer from } F^n_s$
        \State return $L^n_s, H^n, P^n$
\end{algorithmic}
\end{algorithm}

\section{Proposed Method}
\label{sec:method}

\subsection{LW-FedSSL}

Our proposed approach can be integrated with various SSL techniques, and we collectively refer to it as LW-FedSSL.
Fig.~\ref{fig:lwssl} illustrates the local training process for the $n$-th FL client across different stages of LW-FedSSL.
For an encoder $F$ with a total of $S$ layers, the training process can be generally divided into $S$ stages. 
Each stage $s \in [1,S]$ runs for a fixed number of FL communication rounds, $R_s$.
The training process starts with an empty encoder $F_0$ (i.e., with no layers), and each stage $s$ sequentially adds a new layer $L_s$ to $F_{(s-1)}$, forming $F_s$ and increasing the encoder’s depth. 
During stage $s$, only the newly added layer $L_s$ is updated, while all prior layers [$L_1$, \dots, $L_{(s-1)}$] within $F_s$ are kept frozen, serving only for inference. 
Moreover, in each communication round, clients transmit only the active layer $L_s$ (along with the MLP heads, depending on the SSL technique used) to the server for aggregation. 
Likewise, the server only needs to broadcast the aggregated layer $L_s$ back to clients.
The detailed procedure of LW-FedSSL using MoCoV3 as the SSL technique is shown in Algorithms~\ref{alg:lwfedssl_server} and \ref{alg:lwfedssl_client}.
Algorithm~\ref{alg:lwfedssl_server} describes the server-side execution, which manages the incremental training stages $s \in [1, S]$ and the aggregation process.
Each client executes Algorithm~\ref{alg:lwfedssl_client}, performing local SSL training to update the active layer $L_s$.

Dividing the training process into stages that focus on a single layer (or a block of layers) at a time significantly reduces memory and computational demands, making it more suitable for resource-constrained client devices in FL environments. 
Additionally, it lowers both upload and download communication costs for clients, as only the active layers need to be exchanged between the server and clients.
In FL, many clients may lack the resources required for conventional end-to-end training, preventing them from contributing to model training despite having valuable local data.
Our proposed approach can encourage a larger number of client devices with limited computational capacity to participate, leveraging their local data for model training.

In LW-FedSSL, the number of training stages can be mainly determined by the chosen model architecture and the desired level of resource savings. 
It can be easily adjusted to incorporate more than a single layer within each stage, though doing so increases computational and communication demands per stage. 
As a result, there is an inherent trade-off between the number of layers added at each stage and the resources required for training.
This flexibility allows for optimizing the training configuration based on system constraints.

\subsection{Prog-FedSSL}

\begin{figure*}[ht]
    \centering
    \subfloat[FedSSL (end-to-end)]{\includegraphics[width=0.31\linewidth, trim=20pt 0pt 25pt 0pt, clip]{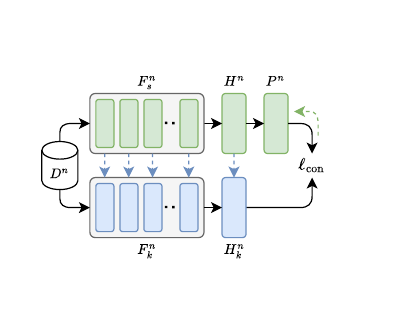} \label{subfig:fedmocov3}} \hspace{5pt}
    \subfloat[LW-FedSSL (layer-wise)]{\includegraphics[width=0.31\linewidth, trim=20pt 0pt 25pt 0pt, clip]{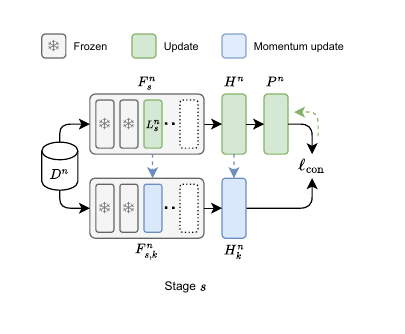} \label{subfig:lwfedmocov3_stage}} \hspace{5pt}
    \subfloat[Prog-FedSSL (progressive)]{\includegraphics[width=0.31\linewidth, trim=20pt 0pt 25pt 0pt, clip]{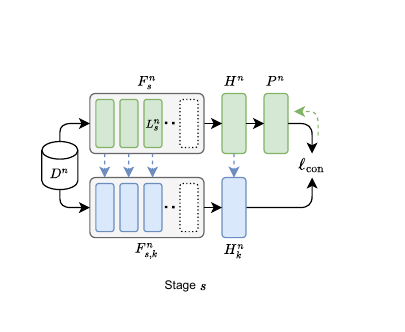} \label{subfig:progfedmocov3_stage}} 
	\caption{Comparison of FedSSL, LW-FedSSL (ours), and Prog-FedSSL (ours) at stage $s$ using MoCoV3 as the SSL backbone.}
	\label{fig:compare_training_methods}
\end{figure*}

Inspired by ProgFed \citep{wang2022progfed}, we also explore progressive training for federated self-supervised learning, denoted as Prog-FedSSL.
The local training process of Prog-FedSSL at stage $s$ is shown in Fig.~\ref{subfig:progfedmocov3_stage}, while a more comprehensive illustration across different stages is provided in Fig.~\ref{fig:prog_fedssl} of \ref{sec:additional_details}.
Similar to layer-wise training, progressive training also starts with an empty encoder $F_0$ and sequentially adds a new layer $L_s$ to $F_{(s-1)}$ at the beginning of each stage $s$.
However, in contrast to layer-wise training, which updates only a single layer $L_s$ at a time, progressive training updates all existing layers (i.e., [$L_1$, \dots, $L_s$]) during each stage $s$. 
In other words, no layers within $F_s$ are frozen at any stage.
This distinction results in higher computational and communication costs for \mbox{Prog-FedSSL} compared to \mbox{LW-FedSSL}. 
Nonetheless, when compared with end-to-end training—where clients must train a full encoder with $S$ layers (i.e., [$L_1$, \dots, $L_S$]) at each round—Prog-FedSSL allows clients to train only $s$ layers $(s \leq S)$. 
Moreover, clients only need to exchange these $s$ layers with the server.
In essence, Prog-FedSSL reduces both computational and communication costs by avoiding the training and exchanging of $(S-s)$ layers at each stage $s$ compared to end-to-end training.
The detailed procedure of Prog-FedSSL using MoCoV3 as the SSL technique is presented  in Algorithms~\ref{alg:progfedssl_server} and \ref{alg:progfedssl_client} of \ref{sec:additional_details}, with Algorithm~\ref{alg:progfedssl_server} describing the server-side execution and Algorithm~\ref{alg:progfedssl_client} describing the client-side execution.

Nevertheless, it is important to recognize that the resource requirements of progressive training gradually increase as the stages progress, eventually matching those of end-to-end training at the final stage (i.e., when $s=S$). 
This growing demand could prevent resource-constrained devices from participating in training, leading to similar limitations faced by the end-to-end approach and resulting in the loss of valuable data from these devices.
Meanwhile, the resource requirements of layer-wise training can be significantly lower at any stage, as it trains only a single layer $L_s$ at a time.
Fig.~\ref{fig:compare_training_methods} illustrates a comparison of FedSSL (end-to-end), LW-FedSSL (layer-wise), and \mbox{Prog-FedSSL} (progressive) at stage $s$.
Additionally, Table~\ref{tab:comparison_characteristics} provides a detailed comparison of their different layer management characteristics.

\begin{table*}[ht]
\caption{Comparison of different layer management characteristics between FedSSL, LW-FedSSL (ours), and Prog-FedSSL (ours).}
\label{tab:comparison_characteristics}
\resizebox{\linewidth}{!}{
\begin{tabular}{@{}l|cccc|ccccccc@{}}
\toprule
\multirow{2}{*}{} & \multicolumn{4}{c|}{Initialization}                                                                       & \multicolumn{7}{c}{At Stage $s \in [1, S]$}                                                                                                                                                                                                                                                                                                              \\ \cmidrule(l){2-12} 
                  & Training    & \begin{tabular}[c]{@{}c@{}}Number \\ of Stages\end{tabular} & Encoder & Layers              & Encoder & Layers              & \begin{tabular}[c]{@{}c@{}}New\\ Layer\end{tabular} & \begin{tabular}[c]{@{}c@{}}Frozen\\ Layers\end{tabular} & \begin{tabular}[c]{@{}c@{}}Trainable\\ Layers\end{tabular} & \begin{tabular}[c]{@{}c@{}}Server to Client\\ (Download)\end{tabular} & \begin{tabular}[c]{@{}c@{}}Client to Server\\ (Upload)\end{tabular} \\ \midrule
FedSSL            & end-to-end  & --                                                          & $F$     & $[L_1, \dots, L_S]$ & $F$     & $[L_1, \dots, L_S]$ & --                                                  & --                                                      & $[L_1, \dots, L_S]$                                        & $[L_1, \dots, L_S]$                                                   & $[L_1, \dots, L_S]$                                                 \\
LW-FedSSL         & layer-wise  & $S$                                                         & $F_0$   & --                  & $F_s$   & $[L_1, \dots, L_s]$ & $L_s$                                               & $[L_1, \dots, L_{(s-1)}]$                                 & $L_s$                                                      & $L_s$                                                                 & $L_s$                                                               \\
Prog-FedSSL       & progressive & $S$                                                         & $F_0$   & --                  & $F_s$   & $[L_1, \dots, L_s]$ & $L_s$                                               & --                                                      & $[L_1, \dots, L_s]$                                        & $[L_1, \dots, L_s]$                                                   & $[L_1, \dots, L_s]$                                                 \\ \bottomrule
\end{tabular}
}
\end{table*}

\section{Experiment}
\label{sec:experiment}

\subsection{Experimental setup}
\label{sec:experimenta_setup}

Unless otherwise stated, we use the following settings as the default in our experiments.

\paragraph{\textbf{Model}}
We use a ViT-Tiny~\citep{dosovitskiy2020image} backbone with 12 transformer blocks as the encoder $F$. 
The projection head $H$ and the prediction head $P$ are implemented as a 3-layer MLP and a 2-layer MLP, respectively. 
For both $H$ and $P$, the hidden layer dimensions are set to 512, and the output dimension is set to 256.

\paragraph{\textbf{Data}}
We use images from the COCO~\cite{lin2014microsoft} dataset to construct the client datasets.
The COCO dataset contains diverse natural scene images, which inherently introduce data heterogeneity. 
These images are uniformly distributed across 10 FL clients. 
To evaluate the performance, we use downstream datasets including \mbox{CIFAR-10/100}~\citep{krizhevsky2009learning}, Tiny ImageNet~\citep{le2015tiny}, and Caltech-101~\citep{li_andreeto_ranzato_perona_2022}. 
For all datasets, we use an image size of $32 \times 32$ with a patch size of 4.

\paragraph{\textbf{Training}}
The number of training stages is set to 12 (i.e., $S=12$).
The total number of FL communication rounds $R$ is set to 180 for all approaches. For our proposed approaches, $R$ is evenly distributed across the stages ${s \in [1,S]}$. 
This results in 15 rounds per stage (i.e., ${R_s = R / S = 15}$).
Each FL client performs local training for 3 epochs using the AdamW optimizer, with a base learning rate of 1.5e-4 and a weight decay of 1e-5. 
The batch size is set to 512, and the learning rate is linearly scaled as \textit{$\text{base\_learning\_rate} \times \text{batch\_size} / \textit{256}$} \citep{goyal2017accurate,krizhevsky2014one}. 
We set the momentum $\mu$ to 0.99 and the temperature $\tau$ to 0.05. 
Data augmentations for view creation include random resized crop, color jitter, grayscale, horizontal flip, Gaussian blur, and solarization.

\paragraph{\textbf{Evaluation}}
After training, the encoder $F$ is retained, while the projection head $H$ and prediction head $P$ are discarded.
A linear classifier head is added to the encoder $F$ and fine-tuned on the downstream datasets for 40 epochs, including a warmup period of 10 epochs.
We set the batch size to 256 and use the AdamW optimizer with a base learning rate of 1e-3 and a weight decay of 1e-5.
A cosine decay schedule is applied to the learning rate during training.
We use the \texttt{RandAugment} function available in the PyTorch~\citep{Paszke_PyTorch_An_Imperative_2019} framework for augmentation.

\begin{table*}[ht]
\centering
\caption{Comparison of resource requirements and performance. For resource requirements, we compare the maximum memory usage (GB), total GFLOPs, and communication costs (download + upload) for a client. Expect for the centralized settings, best values within each group are marked in \textbf{bold}.}
\label{tab:main_performance}
\resizebox{\linewidth}{!}{
\begin{tabular}{l|rrr|ccccc}
\toprule
                      & \multicolumn{3}{c|}{Resource Requirements}                                                                                                                                         & \multicolumn{5}{c}{Accuracy (\%)}                                                                                                                \\ \cmidrule{2-9} 
                      & \multicolumn{1}{c}{Memory (GB)} & \multicolumn{1}{c}{GFLOPs} & \multicolumn{1}{c|}{Comm. (GB)} & CIFAR-10       & CIFAR-100      & \begin{tabular}[c]{@{}c@{}}Tiny\\ ImageNet\end{tabular} & \multicolumn{1}{c|}{Caltech-101}    & Avg            \\ \midrule
Scratch               & \multicolumn{1}{c}{--}                                                    & \multicolumn{1}{c}{--}     & \multicolumn{1}{c|}{--}                                                   & 63.28          & 39.82          & 26.09                                                   & \multicolumn{1}{c|}{22.10}          & 37.82          \\ \midrule
\multicolumn{9}{c}{MoCoV3~\citep{9711302}}                                                                                                                                                                                                                                                                                                                                                      \\ \midrule
MoCoV3~\citep{9711302} (Centralized)                & \multicolumn{1}{c}{--}                                                    & \multicolumn{1}{c}{--}     & \multicolumn{1}{c|}{--}                                                   & 87.22          & 65.97          & 43.08                                                   & \multicolumn{1}{c|}{60.61}          & 64.22          \\
FedMoCoV3             & 8.72 (1.00×)                                                              & 586 (1.00×)                & 8.40 (1.00×)                                                              & 83.95          & 62.84          & \textbf{41.77}                                          & \multicolumn{1}{c|}{54.60}          & 60.79          \\
LW-FedSSL (ours)   & \textbf{2.61 (0.30×)}                                                     & \textbf{139 (0.24×)}       & \textbf{1.66 (0.20×)}                                                     & 83.43          & 61.65          & 40.00                                                   & \multicolumn{1}{c|}{48.33}          & 58.35          \\
Prog-FedSSL (ours) & 8.69 (1.00×)                                                              & 318 (0.54×)                & 5.04 (0.60×)                                                              & \textbf{86.53} & \textbf{65.10} & 40.70                                                   & \multicolumn{1}{c|}{\textbf{61.60}} & \textbf{63.48} \\ \midrule
\multicolumn{9}{c}{BYOL~\citep{grill2020bootstrap}}                                                                                                                                                                                                                                                                                                                                                          \\ \midrule
BYOL~\citep{grill2020bootstrap} (Centralized)                  & \multicolumn{1}{c}{--}                                                    & \multicolumn{1}{c}{--}     & \multicolumn{1}{c|}{--}                                                   & 87.69          & 66.26          & 44.84                                                   & \multicolumn{1}{c|}{62.32}          & 65.28          \\
FedBYOL               & 8.72 (1.00×)                                                              & 586 (1.00×)                & 8.40 (1.00×)                                                              & 83.59          & 62.71          & 42.67                                                   & \multicolumn{1}{c|}{53.89}          & 60.72          \\
FedU~\citep{zhuang2021collaborative}                  & 8.94 (1.03×)                                                              & 586 (1.00×)                & 8.40 (1.00×)                                                              & 83.50          & 62.61          & 42.61                                                   & \multicolumn{1}{c|}{53.88}          & 60.65          \\
FedEMA~\citep{zhuang2022divergence}                & 8.96 (1.03×)                                                              & 586 (1.00×)                & 8.40 (1.00×)                                                              & 81.35          & 60.50          & 41.20                                                   & \multicolumn{1}{c|}{48.26}          & 57.83          \\
LW-FedSSL (ours)     & \textbf{2.61 (0.30×)}                                                     & \textbf{139 (0.24×)}       & \textbf{1.66 (0.20×)}                                                     & 84.20          & 61.29          & 40.56                                                   & \multicolumn{1}{c|}{47.77}          & 58.45          \\
Prog-FedSSL (ours)   & 8.69 (1.00×)                                                              & 318 (0.54×)                & 5.04 (0.60×)                                                              & \textbf{87.51} & \textbf{66.32} & \textbf{43.16}                                          & \multicolumn{1}{c|}{\textbf{62.19}} & \textbf{64.80} \\ \midrule
\multicolumn{9}{c}{SimCLR~\citep{chen2020simple}}                                                                                                                                                                                                                                                                                                                                                        \\ \midrule
SimCLR~\citep{chen2020simple} (Centralized)               & \multicolumn{1}{c}{--}                                                    & \multicolumn{1}{c}{--}     & \multicolumn{1}{c|}{--}                                                   & 87.49          & 65.64          & 43.54                                                   & \multicolumn{1}{c|}{61.12}          & 64.45          \\
FedSimCLR             & 14.39 (1.00×)                                                             & 1169 (1.00×)               & 8.03 (1.00×)                                                              & 84.38          & 62.41          & 42.49                                                   & \multicolumn{1}{c|}{53.45}          & 60.68          \\
FedCA~\citep{zhang2023federated}                 & 16.37 (1.14×)                                                             & 1299 (1.11×)               & 30.69 (3.82×)                                                             & 85.30          & 63.68          & \textbf{43.04}                                          & \multicolumn{1}{c|}{57.75}          & 62.44          \\
LW-FedSSL (ours)   & \textbf{8.59 (0.60×)}                                                     & \textbf{278 (0.24×)}       & \textbf{1.29 (0.16×)}                                                     & 84.27          & 61.74          & 40.70                                                   & \multicolumn{1}{c|}{48.25}          & 58.74          \\
Prog-FedSSL (ours) & 14.36 (1.00×)                                                             & 635 (0.54×)                & 4.68 (0.58×)                                                              & \textbf{86.38} & \textbf{63.77} & 41.60                                                   & \multicolumn{1}{c|}{\textbf{62.08}} & \textbf{63.46} \\ \bottomrule
\end{tabular}
}
\end{table*}

\begin{figure*}[ht]
    \centering
    \subfloat[Memory.]{\includegraphics[width=0.29\linewidth]{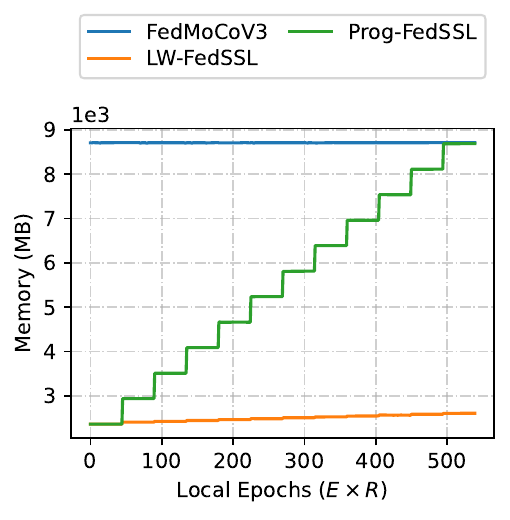} \label{subfig:mem_mocov3}} \hspace{2em}
    \subfloat[FLOPs.]{\includegraphics[width=0.29\linewidth]{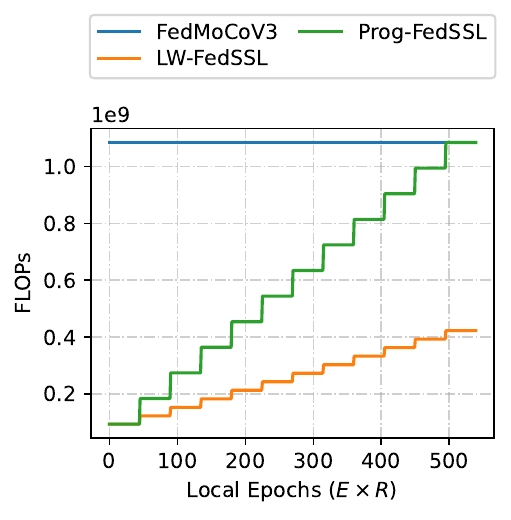} \label{subfig:flops_mocov3}} \hspace{2em}
    \subfloat[Communication.]{\includegraphics[width=0.29\linewidth]{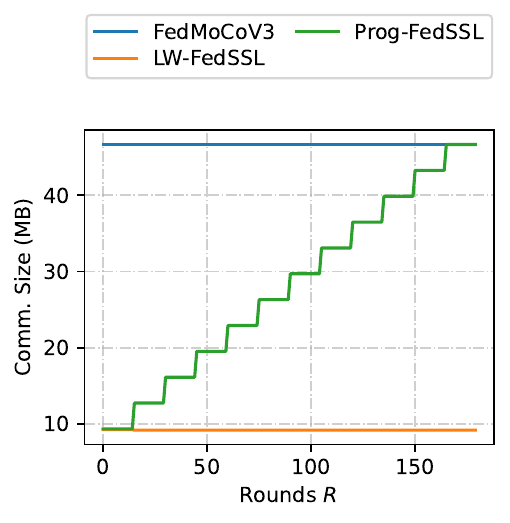} \label{subfig:comm_cost_mocov3}}
	\caption{Computational and communication resources required for a client (a) Memory usage. (b) FLOPSs consumption. (c) Communication cost.}
	\label{fig:resource_plot}
\end{figure*}

\subsection{Performance and Resource Comparison}
\label{sec:performance_and_resource}

Table~\ref{tab:main_performance} presents a comparison of the resource requirements and performance across different training approaches.\footnote{We report mean per-class accuracy for the Caltech-101 dataset.}
For resource requirements, we compare the computational and communication resources needed for a single FL client.
Additionally, we demonstrate that our proposed methods, LW-FedSSL and Prog-FedSSL, can be seamlessly integrated with various SSL techniques. 
As shown in Table~\ref{tab:main_performance}, we integrate MoCoV3~\citep{9711302}, BYOL~\citep{grill2020bootstrap}, and SimCLR~\citep{chen2020simple} with our methods.
For clarity, we group similar methods based on the SSL technique used.
Fig.~\ref{fig:resource_plot} plots the resource consumption  for a client, illustrating computational resources (i.e., memory and FLOPs) across local epochs and communication cost (download + upload) across FL rounds.
For FLOPs calculation, we consider only a single input~sample.

The results in Table~\ref{tab:main_performance} show that LW-FedSSL significantly reduces resource requirements.
Specifically, when using MoCoV3, LW-FedSSL achieves a $3.34 \times$ reduction in memory usage, a $4.20 \times$ fewer computational operations (GFLOPs), and a $5.07 \times$ reduction in communication costs compared to conventional FedMoCoV3.
Additionally, Table~\ref{tab:main_performance} demonstrates that LW-FedSSL also maintains comparable performance to its end-to-end training counterparts.
While Prog-FedSSL is not as resource-efficient as LW-FedSSL, it still consumes fewer GFLOPs and communication resources compared to FedMoCoV3.
Furthermore, Prog-FedSSL achieves better performance in most cases, suggesting that progressive training can also benefit self-supervised learning, aligning with the findings for supervised learning in \citep{wang2022progfed}.
Fig.~\ref{fig:resource_plot} illustrates that the resource consumption for LW-FedSSL remains relatively flat across local epochs and FL rounds due to the focus on training a single layer at each stage.
Meanwhile, the resource consumption for Prog-FedSSL gradually increases as the number of trainable layers grows with each stage, eventually matching that of conventional FedMoCoV3.

\subsection{Worst-case Communication Scenario for LW-FedSSL}

By default, we assume that clients only need to download the layer $L_s$ at each FL round for LW-FedSSL.
This assumption holds only if all clients participate in every FL round without dropping out or no new clients join.
Otherwise, new clients or those that previously dropped out would need to download the latest model $F_s$ (i.e., $[L_1, \dots, L_s]$) instead of just $L_s$.
Therefore, we compare the worst-case communication cost for LW-FedSSL in Table~\ref{tab:worst_comm_lwfedssl}, where a client downloads $F_s$ at every FL round.
Note that client still needs to upload only $L_s$ after local training. 
The results show that even in the worst-case scenario, LW-FedSSL significantly reduces communication costs compared to end-to-end FedMoCoV3.

\begin{table}[ht]
\centering
\caption{Communication cost for the worst-case scenario of LW-FedSSL.}
\label{tab:worst_comm_lwfedssl}
\resizebox{0.26\linewidth}{!}{
\begin{tabular}{l|c}
\toprule
                   & Comm. (GB)            \\ \midrule
FedMoCoV3          & 8.40 (1.00×)          \\
LW-FedSSL (ours)   & \textbf{3.35 (0.40×)} \\
Prog-FedSSL (ours) & 5.04 (0.60×)          \\ \bottomrule
\end{tabular}
}
\end{table}

\begin{table*}[ht]
\centering
\caption{Resource requirements and performance comparison for different numbers of layers added per stage in LW-FedSSL and Prog-FedSSL.}
\label{tab:stage_num_layers}
\resizebox{\linewidth}{!}{
\begin{tabular}{@{}c|ccc|ccc|ccccc@{}}
\toprule
                             & \multicolumn{3}{c|}{Stage Settings}                                                                                                                                                       & \multicolumn{3}{c|}{Resource Requirements}                           & \multicolumn{5}{c}{Accuracy (\%)}                                                                                                                 \\ \cmidrule(l){2-12} 
                             & \begin{tabular}[c]{@{}c@{}}Layers Added\\ per Stage\end{tabular} & \begin{tabular}[c]{@{}c@{}}Stages\\ $S$\end{tabular} & \begin{tabular}[c]{@{}c@{}}Rounds per \\ Stage $R_s$\end{tabular} & Memory (GB)           & GFLOPs               & Comm. (GB)            & CIFAR-10       & CIFAR-100      & \begin{tabular}[c]{@{}c@{}}Tiny\\ ImageNet\end{tabular} & \multicolumn{1}{c|}{Caltech-101}    & Avg            \\ \midrule
FedMoCoV3                    & --                                                             & --                                                   & --                                                                & 8.72 (1.00×)          & 586 (1.00×)          & 8.40 (1.00×)          & 83.95          & 62.84          & 41.77                                                   & \multicolumn{1}{c|}{54.60}          & 60.79          \\ \midrule
\multirow{5}{*}{\begin{tabular}[c]{@{}c@{}}LW-FedSSL\\ (ours)\end{tabular}}   & 1                                                              & 12                                                   & 15                                                                & \textbf{2.61 (0.30×)} & \textbf{139 (0.24×)} & \textbf{1.66 (0.20×)} & 83.43          & 61.65          & 40.00                                                   & \multicolumn{1}{c|}{48.33}          & 58.35          \\
                             & 2                                                              & 6                                                    & 30                                                                & 3.12 (0.36×)          & 180 (0.31×)          & 2.27 (0.27×)          & 84.85          & 62.23          & 40.11                                                   & \multicolumn{1}{c|}{56.19}          & 60.84          \\
                             & 3                                                              & 4                                                    & 45                                                                & 3.67 (0.42×)          & 220 (0.38×)          & 2.88 (0.34×)          & 85.32          & 63.07          & 40.28                                                   & \multicolumn{1}{c|}{58.48}          & 61.79          \\
                             & 4                                                              & 3                                                    & 60                                                                & 4.23 (0.48×)          & 261 (0.45×)          & 3.49 (0.42×)          & 85.54          & 64.13          & 40.71                                                   & \multicolumn{1}{c|}{58.64}          & 62.26          \\
                             & 6                                                              & 2                                                    & 90                                                                & 5.34 (0.61×)          & 342 (0.58×)          & 4.71 (0.56×)          & 86.58          & 66.18          & 42.79                                                   & \multicolumn{1}{c|}{62.72}          & 64.57          \\ \midrule
\multirow{5}{*}{\begin{tabular}[c]{@{}c@{}}Prog-FedSSL\\ (ours)\end{tabular}} & 1                                                              & 12                                                   & 15                                                                & 8.69 (1.00×)          & 318 (0.54×)          & 5.04 (0.60×)          & 86.53          & 65.10          & 40.70                                                   & \multicolumn{1}{c|}{61.60}          & 63.48          \\
                             & 2                                                              & 6                                                    & 30                                                                & 8.69 (1.00×)          & 342 (0.58×)          & 5.35 (0.64×)          & 86.57          & 65.39          & 42.00                                                   & \multicolumn{1}{c|}{62.84}          & 64.20          \\
                             & 3                                                              & 4                                                    & 45                                                                & 8.69 (1.00×)          & 367 (0.63×)          & 5.65 (0.67×)          & 86.61          & 65.73          & 42.13                                                   & \multicolumn{1}{c|}{63.52}          & 64.50          \\
                             & 4                                                              & 3                                                    & 60                                                                & 8.69 (1.00×)          & 391 (0.67×)          & 5.96 (0.71×)          & 87.33          & 65.80          & 43.12                                                   & \multicolumn{1}{c|}{62.56}          & 64.70          \\
                             & 6                                                              & 2                                                    & 90                                                                & 8.69 (1.00×)          & 440 (0.75×)          & 6.57 (0.78×)          & \textbf{88.08} & \textbf{66.78} & \textbf{43.80}                                          & \multicolumn{1}{c|}{\textbf{64.21}} & \textbf{65.72} \\ \bottomrule
\end{tabular}
}
\end{table*}

\subsection{Number of Layers Added per Stage}

Both LW-FedSSL and Prog-FedSSL allow each stage to be adjusted to incorporate multiple new layers instead of just a single layer.
However, increasing the number of layers per stage also raises resource requirements, including memory, computational cost, and communication overhead.
The number of layers added per stage presents a trade-off between training efficiency and resource constraints, which can be adjusted based on the resource capacity of the targeted FL clients.
Table~\ref{tab:stage_num_layers} compares different settings for LW-FedSSL and Prog-FedSSL, where varying numbers of new layers are added at each stage.
In all cases, the total number of training rounds $R$ is kept constant at 180.
As more layers are added per stage, the number of stages decreases, and resource requirements increase, but this is accompanied by an improvement in model performance.

\subsection{ResNet-18}

To demonstrate that our proposed approaches can also work well with different model architectures, we use ResNet-18~\citep{he2016deep} as the encoder $F$ in Table~\ref{tab:resnet}.
We set the number of stages to $S=4$, with each stage sequentially adding a new ResNet block to the encoder.
The total number of FL rounds is set to $R=120$, with each stage $s$ allocated $R_s=30$ rounds.
Whenever a new ResNet block is added and the output dimension of the encoder changes, we add a temporary linear layer to reshape the output to 512. 
This reshaping layer is discarded at the end of each stage.
The results in Table~\ref{tab:resnet} demonstrate that our approaches also perform well with ResNet-18.

\begin{table*}[ht]
\centering
\caption{Comparison of resource requirements and performance with ResNet-18 as the encoder.}
\label{tab:resnet}
\resizebox{0.9\linewidth}{!}{
\begin{tabular}{@{}l|rrr|ccccc@{}}
\toprule
                   & \multicolumn{3}{c|}{Resource Requirements}                                                     & \multicolumn{5}{c}{Accuracy (\%)}                                                                                                                \\ \cmidrule(l){2-9} 
                   & \multicolumn{1}{c}{Memory (GB)} & \multicolumn{1}{c}{GFLOPs} & \multicolumn{1}{c|}{Comm. (GB)} & CIFAR-10       & CIFAR-100      & \begin{tabular}[c]{@{}c@{}}Tiny\\ ImageNet\end{tabular} & \multicolumn{1}{c|}{Caltech-101}    & Avg            \\ \midrule
FedMoCoV3          & 1.83 (1.00×)                    & 41 (1.00×)                 & 11.09 (1.00×)                   & 85.20          & 57.57          & 33.71                                                   & \multicolumn{1}{c|}{61.58}          & 59.52          \\
LW-FedSSL (ours)   & \textbf{1.59 (0.87×)}           & \textbf{17 (0.40×)}        & \textbf{3.41 (0.31×)}           & \textbf{85.64} & \textbf{57.84} & 34.40                                                   & \multicolumn{1}{c|}{61.84}          & 59.93          \\
Prog-FedSSL (ours) & 1.81 (0.99×)                    & 27 (0.67×)                 & 4.24 (0.38×)                    & 85.57          & 57.44          & \textbf{34.63}                                          & \multicolumn{1}{c|}{\textbf{63.59}} & \textbf{60.31} \\ \bottomrule
\end{tabular}
}
\end{table*}

\subsection{Integration of Additional Mechanisms}

In both LW-FedSSL and Prog-FedSSL, each stage can be viewed as an independent FL process. 
At the beginning of each stage, some layers retain weights from the previous stage, but apart from that, the FL process within each stage remains independent. 
This modularity allows easy integration of existing FL mechanisms into LW-FedSSL and Prog-FedSSL.
To demonstrate this flexibility, we incorporate depth dropout (DD) \citep{pengfei2023towards}, a technique that selectively drops frozen layers during training; alignment using the proximal term (PA) \citep{li2020federated}, which constrains local updates to prevent drastic weight divergence across clients; and a similar alignment mechanism using representation (RA) \citep{Li_2021_CVPR}.
Table~\ref{tab:integration_compare} presents the evaluation results, including resource requirements and downstream performance across multiple datasets.

\begin{table*}[ht]
\centering
\caption{Resource requirements and performance comparison when integrating Depth Dropout (DD), Proximal Alignment (PA), and Representation Alignment (RA) into LW-FedSSL and Prog-FedSSL.}
\label{tab:integration_compare}
\resizebox{0.95\linewidth}{!}{
\begin{tabular}{@{}l|ccc|ccccc@{}}
\toprule
                 & \multicolumn{3}{c|}{Resource Requirements}                           & \multicolumn{5}{c}{Accuracy (\%)}                                                                                                                 \\ \cmidrule(l){2-9} 
                 & Memory (GB)           & GFLOPs               & Comm. (GB)            & CIFAR-10       & CIFAR-100      & \begin{tabular}[c]{@{}c@{}}Tiny\\ ImageNet\end{tabular} & \multicolumn{1}{c|}{Caltech-101}    & Avg            \\ \midrule
LW-FedSSL        & 2.61 (1.00×)          & 139 (1.00×)          & \textbf{1.66 (1.00×)} & 83.43          & 61.65          & 40.00                                                   & \multicolumn{1}{c|}{48.33}          & 58.35          \\
LW-FedSSL + DD \cite{pengfei2023towards}   & \textbf{2.49 (0.95×)} & \textbf{111 (0.80×)} & \textbf{1.66 (1.00×)} & 84.10          & 60.78          & 40.18                                                   & \multicolumn{1}{c|}{\textbf{48.90}} & \textbf{58.49} \\
LW-FedSSL + PA \cite{li2020federated}   & 2.63 (1.01×)          & 139 (1.00×)          & \textbf{1.66 (1.00×)} & \textbf{84.46} & \textbf{61.74} & \textbf{40.72}                                          & \multicolumn{1}{c|}{44.31}          & 57.81          \\
LW-FedSSL + RA \cite{Li_2021_CVPR}   & 2.86 (1.10×)          & 318 (2.29×)          & \textbf{1.66 (1.00×)} & 84.31          & 61.69          & 40.18                                                   & \multicolumn{1}{c|}{45.29}          & 57.87          \\ \midrule
Prog-FedSSL      & 8.69 (1.00×)          & 318 (1.00×)          & 5.04 (1.00×)          & 86.53          & 65.10          & 40.70                                                   & \multicolumn{1}{c|}{61.60}          & 63.48          \\
Prog-FedSSL + DD \cite{pengfei2023towards} & \textbf{5.24 (0.60×)} & \textbf{233 (0.73×)} & \textbf{3.97 (0.79×)} & 86.69          & 64.41          & 41.59                                                   & \multicolumn{1}{c|}{\textbf{63.46}} & 64.04          \\
Prog-FedSSL + PA \cite{li2020federated} & 8.72 (1.00×)          & 318 (1.00×)          & 5.04 (1.00×)          & 84.71          & 62.09          & 38.65                                                   & \multicolumn{1}{c|}{56.35}          & 60.45          \\
Prog-FedSSL + RA \cite{Li_2021_CVPR} & 9.00 (1.04×)          & 496 (1.56×)          & 5.04 (1.00×)          & \textbf{87.05} & \textbf{66.05} & \textbf{42.37}                                          & \multicolumn{1}{c|}{62.59}          & \textbf{64.52} \\ \bottomrule
\end{tabular}
}
\end{table*}

\subsection{Partial Client Participation}

In FL environments, it is common for only a subset of clients to participate in the training at each communication round. 
Clients may drop out of the training due to various constraints, such as unstable network connections or power limitations.
In Table~\ref{tab:client_participation_flat}, we set the total number of clients to 100 and randomly select 25 or 50 clients to participate in each FL round.
Furthermore, in Table~\ref{tab:client_participation_prob}, we use another setting where we assign a dropout probability to each client, using dropout rates of 25\% or 50\%.

\begin{figure}[H]
    \centering
    \includegraphics[width=0.55\linewidth]{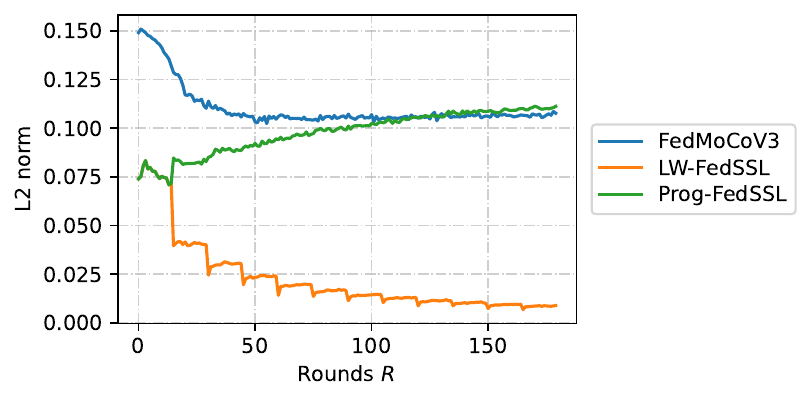}
    \caption{Distance between local model weights. At the end of each FL round, we randomly select a client and measure the L2 norm between its local model weights and those of all other clients, plotting the average.}    
    \label{fig:divergence_mocov3}
\end{figure}

Interestingly, we observe that LW-FedSSL consistently outperforms the  end-to-end FedMoCoV3 under both settings.
To investigate this, in Fig.~\ref{fig:divergence_mocov3}, we measure the L2 norm distance between local model weights at the end of each FL round. We use the setting where 25 clients are selected per round.
To measure the distance, we randomly select a client and compute the average L2 norm between its local model weights and those of all other clients.
Fig.~\ref{fig:divergence_mocov3} indicates that gradually adding new trainable layers, stage by stage, helps limit the divergence between local models. 
We believe this can have a positive impact on the aggregation process, especially with a large number of clients, leading to improved performance.

\begin{table}[ht]
\centering
\caption{Performance comparison under partial client participation, where a subset of clients is selected at each FL round. The setting considers 25 or 50 out of 100 clients participating in each round.}
\label{tab:client_participation_flat}
\resizebox{0.6\linewidth}{!}{
\begin{tabular}{@{}lccccc@{}}
\toprule
\multicolumn{1}{l|}{\multirow{2}{*}{}}  & \multicolumn{5}{c}{Accuracy (\%)}                                                                                                                \\ \cmidrule(l){2-6} 
\multicolumn{1}{l|}{}                   & CIFAR-10       & CIFAR-100      & \begin{tabular}[c]{@{}c@{}}Tiny\\ ImageNet\end{tabular} & \multicolumn{1}{c|}{Caltech-101}    & Avg            \\ \midrule
\multicolumn{6}{c}{Number of Participating Clients: $25 / 100$}                                                                                                                            \\ \midrule
\multicolumn{1}{l|}{FedMoCoV3}          & 71.85          & 49.19          & 32.98                                                   & \multicolumn{1}{c|}{33.95}          & 46.99          \\
\multicolumn{1}{l|}{LW-FedSSL (ours)}   & 84.33          & 61.96          & 39.46                                                   & \multicolumn{1}{c|}{43.80}          & 57.39          \\
\multicolumn{1}{l|}{Prog-FedSSL (ours)} & \textbf{85.18} & \textbf{62.80} & \textbf{40.05}                                          & \multicolumn{1}{c|}{\textbf{55.88}} & \textbf{60.98} \\ \midrule
\multicolumn{6}{c}{Number of Participating Clients: $50 / 100$}                                                                                                                            \\ \midrule
\multicolumn{1}{l|}{FedMoCoV3}          & 72.73          & 49.62          & 33.58                                                   & \multicolumn{1}{c|}{34.67}          & 47.65          \\
\multicolumn{1}{l|}{LW-FedSSL (ours)}   & 84.23          & 62.17          & 41.17                                                   & \multicolumn{1}{c|}{44.28}          & 57.96          \\
\multicolumn{1}{l|}{Prog-FedSSL (ours)} & \textbf{84.46} & \textbf{62.87} & \textbf{40.19}                                          & \multicolumn{1}{c|}{\textbf{56.96}} & \textbf{61.12} \\ \bottomrule
\end{tabular}
}
\end{table}

\begin{table}[ht]
\centering
\caption{Performance comparison under partial client participation, where each client is assigned a dropout probability of 25\% or 50\%. The total number of clients is set to 100.}
\label{tab:client_participation_prob}
\resizebox{0.6\linewidth}{!}{
\begin{tabular}{@{}lccccc@{}}
\toprule
\multicolumn{1}{l|}{\multirow{2}{*}{}}  & \multicolumn{5}{c}{Accuracy (\%)}                                                                                                                \\ \cmidrule(l){2-6} 
\multicolumn{1}{l|}{}                   & CIFAR-10       & CIFAR-100      & \begin{tabular}[c]{@{}c@{}}Tiny\\ ImageNet\end{tabular} & \multicolumn{1}{c|}{Caltech-101}    & Avg            \\ \midrule
\multicolumn{6}{c}{Client Dropout Probability: 25 \%}                                                                                                                                      \\ \midrule
\multicolumn{1}{l|}{FedMoCoV3}          & 72.38          & 49.49          & 33.48                                                   & \multicolumn{1}{c|}{34.38}          & 47.43          \\
\multicolumn{1}{l|}{LW-FedSSL (ours)}   & 84.33          & 62.12          & 40.32                                                   & \multicolumn{1}{c|}{43.19}          & 57.49          \\
\multicolumn{1}{l|}{Prog-FedSSL (ours)} & \textbf{85.11} & \textbf{62.73} & \textbf{40.34}                                          & \multicolumn{1}{c|}{\textbf{54.69}} & \textbf{60.72} \\ \midrule
\multicolumn{6}{c}{Client Dropout Probability: 50 \%}                                                                                                                                      \\ \midrule
\multicolumn{1}{l|}{FedMoCoV3}          & 73.08          & 49.67          & 33.19                                                   & \multicolumn{1}{c|}{35.14}          & 47.77          \\
\multicolumn{1}{l|}{LW-FedSSL (ours)}   & 83.93          & 61.66          & 40.06                                                   & \multicolumn{1}{c|}{44.77}          & 57.61          \\
\multicolumn{1}{l|}{Prog-FedSSL (ours)} & \textbf{84.78} & \textbf{63.08} & \textbf{40.67}                                          & \multicolumn{1}{c|}{\textbf{54.02}} & \textbf{60.64} \\ \bottomrule
\end{tabular}
}
\end{table}

\subsection{Medical Data}
\label{sec:medmnist_main}

\begin{figure*}[ht]
    \centering
    \subfloat[PathMNIST]{\includegraphics[width=0.22\linewidth]{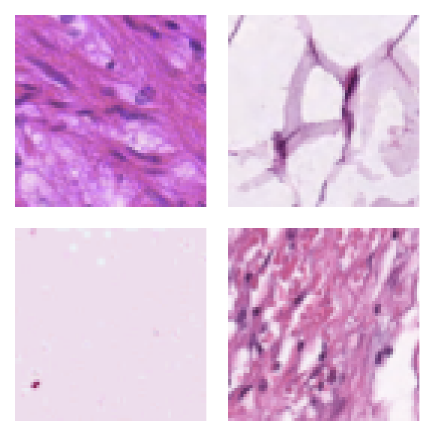} \label{subfig:pathmnist_grid}} \hspace{0.4em}
    \subfloat[BloodMNIST]{\includegraphics[width=0.22\linewidth]{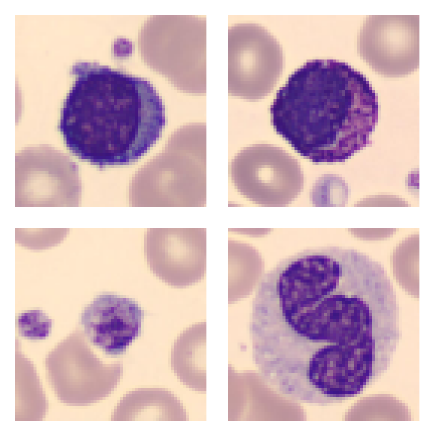} \label{subfig:bloodmnist_grid}}
    \hspace{0.4em}
    \subfloat[DermaMNIST]{\includegraphics[width=0.22\linewidth]{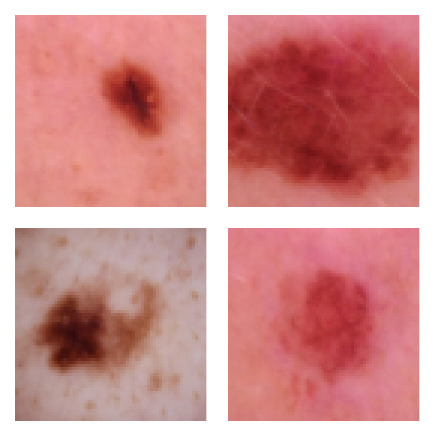} \label{subfig:dermamnist_grid}}
    \hspace{0.4em}
    \subfloat[OrganAMNIST]{\includegraphics[width=0.22\linewidth]{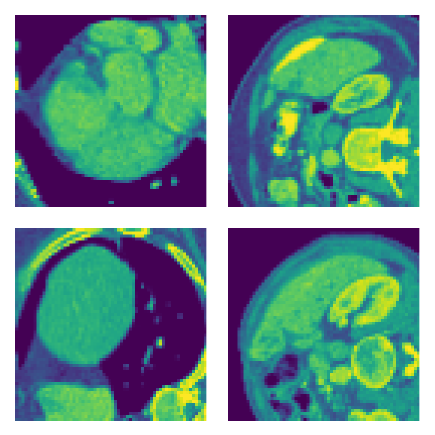} \label{subfig:organamnist_grid}} 
	\caption{Sample images for (a) PathMNIST, (b) BloodMNIST, (c) DermaMNIST, and (d) OrganAMNIST from the MedMNIST collection. Each dataset represents different biomedical imaging modalities used for classification tasks.}
	\label{fig:medmnist_samples}
\end{figure*}

We evaluate our proposed approaches on MedMNIST~\citep{medmnistv2}, a collection of diverse biomedical image datasets. 
Specifically, we use four subsets within MedMNIST: PathMNIST, BloodMNIST, DermaMNIST, and OrganAMNIST. 
Sample images from each dataset are shown in Fig.~\ref{fig:medmnist_samples}.
Each dataset contains three splits: train, validation, and test. 
We use the train split to construct the local datasets for FL clients, the validation split as the downstream dataset for fine-tuning, and the test split for final evaluation.
To introduce data heterogeneity, we partition the data among clients using the Dirichlet distribution~\citep{ferguson1973bayesian} with a concentration parameter of $\beta=5$.
We use an image size of $64 \times 64$  and a batch size of 128 for all experiments. 
For DermaMNIST, we use a 3-layer classifier head with a hidden dimension of 256 and fine-tune for 100 epochs. 
All other settings follow the default configuration.
In Table~\ref{tab:medmnist_resource}, we present resource requirements only for BloodMNIST, as the experimental settings—such as batch size, input dimensions, and model architecture—are consistent across all datasets, resulting in similar resource requirements.
In Table~\ref{tab:medmnist_performance}, we evaluate the performance of our proposed methods on the medical datasets.
For all datasets, we report mean per-class accuracy.

\begin{table}[ht]
\centering
\caption{Resource requirements for the BloodMNIST dataset.}
\label{tab:medmnist_resource}
\resizebox{0.5\linewidth}{!}{
\begin{tabular}{@{}l|rrr@{}}
\toprule
                   & \multicolumn{3}{c}{Resource Requirements (BloodMNIST)}                                                                                                                            \\ \cmidrule(l){2-4} 
                   & \multicolumn{1}{c}{\begin{tabular}[c]{@{}c@{}}Memory\\ (GB)\end{tabular}} & \multicolumn{1}{c}{GFLOPs} & \multicolumn{1}{c}{\begin{tabular}[c]{@{}c@{}}Comm.\\ (GB)\end{tabular}} \\ \midrule
FedMoCoV3          & 11.48 (1.00×)                                                             & 2705 (1.00×)               & 8.45 (1.00×)                                                             \\
LW-FedSSL (ours)   & \textbf{3.10 (0.27×)}                                                     & \textbf{640 (0.24×)}       & \textbf{1.66 (0.20×)}                                                    \\
Prog-FedSSL (ours) & 11.46 (1.00×)                                                             & 1468 (0.54×)               & 5.09 (0.60×)                                                             \\ \bottomrule
\end{tabular}
}
\end{table}

\begin{table}[ht]
\centering
\caption{Performance evaluation across medical datasets, including PathMNIST, BloodMNIST, DermaMNIST, and OrganAMNIST, from the MedMNIST collection.}
\label{tab:medmnist_performance}
\resizebox{0.5\linewidth}{!}{
\begin{tabular}{@{}lcccc@{}}
\toprule
\multicolumn{1}{l|}{}                   & Precision      & Recall         & F1             & Accuracy       \\ \midrule
\multicolumn{5}{c}{PathMNIST}                                                                               \\ \midrule
\multicolumn{1}{l|}{FedMoCoV3}          & 86.53          & 86.13          & 85.04          & 85.92          \\
\multicolumn{1}{l|}{LW-FedSSL (ours)}   & 88.13          & 87.92          & 87.22          & 87.24          \\
\multicolumn{1}{l|}{Prog-FedSSL (ours)} & \textbf{90.59} & \textbf{90.25} & \textbf{89.80} & \textbf{90.03} \\ \midrule
\multicolumn{5}{c}{BloodMNIST}                                                                              \\ \midrule
\multicolumn{1}{l|}{FedMoCoV3}          & 91.18          & 90.84          & 90.68          & 90.82          \\
\multicolumn{1}{l|}{LW-FedSSL (ours)}   & 92.90          & 93.15          & 92.76          & 93.20          \\
\multicolumn{1}{l|}{Prog-FedSSL (ours)} & \textbf{94.12} & \textbf{93.73} & \textbf{93.77} & \textbf{93.62} \\ \midrule

\multicolumn{5}{c}{DermaMNIST}                                                                             \\ \midrule
\multicolumn{1}{l|}{FedMoCoV3}          & 50.63          & 48.27          & 48.27          & 45.76          \\
\multicolumn{1}{l|}{LW-FedSSL (ours)}   & 51.33          & 47.55          & 48.15          & 46.06          \\
\multicolumn{1}{l|}{Prog-FedSSL (ours)} & \textbf{51.73} & \textbf{49.38} & \textbf{49.15} & \textbf{47.72} \\ \midrule

\multicolumn{5}{c}{OrganAMNIST}                                                                             \\ \midrule
\multicolumn{1}{l|}{FedMoCoV3}          & 77.28          & 74.40          & 74.31          & 74.19          \\
\multicolumn{1}{l|}{LW-FedSSL (ours)}   & 77.05          & 74.27          & 74.10          & 74.32          \\
\multicolumn{1}{l|}{Prog-FedSSL (ours)} & \textbf{78.44} & \textbf{76.61} & \textbf{76.03} & \textbf{76.59} \\ \bottomrule
\end{tabular}
}
\end{table}

\subsection{Impact of Batch Size Variation}

\begin{figure}[ht]
    \centering
    \includegraphics[width=0.5\linewidth]{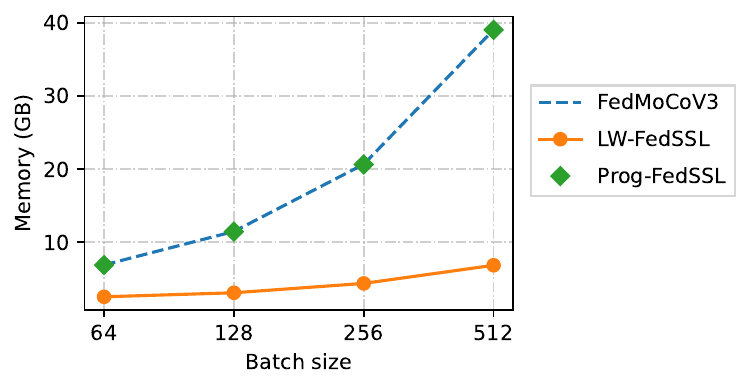}
    \caption{Peak memory usage across different batch sizes on the BloodMNIST dataset.}
    \label{fig:batch_size_gpu_mem}
\end{figure}

We evaluate the impact of batch size on memory consumption using the BloodMNIST dataset.
We consider batch sizes of 64, 128, 256, and 512 while all other settings remain the same as described in Section~\ref{sec:medmnist_main}.
Fig.~\ref{fig:batch_size_gpu_mem} shows the peak memory usage across different batch sizes, while Table~\ref{tab:medmnist_batch_size} presents the evaluation results.
We only report memory usage in Fig.~\ref{fig:batch_size_gpu_mem}, as the FLOPs and communication costs across different batch sizes remain similar to those presented in Table~\ref{tab:medmnist_resource}.\footnote{This is because we consider only a single input sample for FLOPs calculations, and communication cost is independent of the training batch size.}
Given that \mbox{Prog-FedSSL} does not involve frozen layers, its peak memory requirement—which occurs at the final stage $S$—is equivalent to that of end-to-end FedMoCoV3.
Peak memory requirements for both FedMoCoV3 and \mbox{Prog-FedSSL} sharply rise as the batch size grows, while those of \mbox{LW-FedSSL} remain relatively flat. 
Many SSL approaches that rely on contrastive loss require large batch sizes to leverage a greater number of negative samples \citep{chen2020simple}.
The low-memory footprint of LW-FedSSL can accommodate large-batch training, making it more scalable in resource-constrained environments.

\begin{table}[ht]
\centering
\caption{Evaluation on the BloodMNIST dataset across different batch sizes.}
\label{tab:medmnist_batch_size}
\resizebox{0.5\linewidth}{!}{
\begin{tabular}{@{}lcccc@{}}
\toprule
\multicolumn{1}{c|}{}                   & Precision      & Recall         & F1             & Accuracy       \\ \midrule
\multicolumn{5}{c}{Batch Size: 64}                                                                          \\ \midrule
\multicolumn{1}{l|}{FedMoCoV3}          & 92.28          & 91.56          & 91.71          & 91.29          \\
\multicolumn{1}{l|}{LW-FedSSL (ours)}   & 92.66          & 92.85          & 92.61          & 92.61          \\
\multicolumn{1}{l|}{Prog-FedSSL (ours)} & \textbf{93.80} & \textbf{94.25} & \textbf{93.87} & \textbf{94.23} \\ \midrule
\multicolumn{5}{c}{Batch Size: 128}                                                                         \\ \midrule
\multicolumn{1}{l|}{FedMoCoV3}          & 91.18          & 90.84          & 90.68          & 90.82          \\
\multicolumn{1}{l|}{LW-FedSSL (ours)}   & 92.90          & 93.15          & 92.76          & 93.20          \\
\multicolumn{1}{l|}{Prog-FedSSL (ours)} & \textbf{94.12} & \textbf{93.73} & \textbf{93.77} & \textbf{93.62} \\ \midrule
\multicolumn{5}{c}{Batch Size: 256}                                                                         \\ \midrule
\multicolumn{1}{l|}{FedMoCoV3}          & 91.48          & 91.78          & 91.35          & 91.65          \\
\multicolumn{1}{l|}{LW-FedSSL (ours)}   & 92.33          & 92.74          & 92.31          & 92.58          \\
\multicolumn{1}{l|}{Prog-FedSSL (ours)} & \textbf{94.67} & \textbf{94.72} & \textbf{94.54} & \textbf{94.68} \\ \midrule
\multicolumn{5}{c}{Batch Size: 512}                                                                         \\ \midrule
\multicolumn{1}{l|}{FedMoCoV3}          & 90.10          & 90.48          & 90.02          & 90.13          \\
\multicolumn{1}{l|}{LW-FedSSL (ours)}   & 91.95          & 92.25          & 91.86          & 92.01          \\
\multicolumn{1}{l|}{Prog-FedSSL (ours)} & \textbf{93.80} & \textbf{93.91} & \textbf{93.65} & \textbf{93.89} \\ \bottomrule
\end{tabular}
}
\end{table}

\subsection{Impact of Input Dimensions}

In Table~\ref{tab:medmnist_input_size}, we evaluate our proposed approaches using different input image sizes and corresponding patch sizes on the BloodMNIST dataset.
The choice of input size and patch size can significantly impact computational resource requirements.
Larger input dimensions demand increased memory and FLOPs for processing. 
Meanwhile, larger patch sizes reduce the number of patches but may affect feature granularity, potentially influencing model performance.
We compare three different input configurations: (1) a $64 \times 64$ image with $4 \times 4$ patches, (2) a $128 \times 128$ image with $8 \times 8$ patches, and (3) a $224 \times 224$ image with $16 \times 16$ patches.
All other experimental settings are kept as described in Section~\ref{sec:medmnist_main}.
The results indicate that LW-FedSSL consistently maintains its resource efficiency across different input dimensions while achieving strong performance.
Meanwhile, Prog-FedSSL achieves the best overall performance across all input sizes.

\subsection{Data Heterogeneity}
\label{sec:data_heterogeneity}

Data heterogeneity is one of the main challenges in an FL environment.
The concentration parameter~$\beta$ in the Dirichlet distribution can be used to  determine the strength of data heterogeneity, with a lower $\beta$ value indicating a higher degree of heterogeneity. 
In Table~\ref{tab:medmnist_beta}, we conduct experiments using different levels of data heterogeneity by setting $\beta$ values to 50, 5, and 0.5.
\footnote{The resulting client data distributions are visualized in Fig.~\ref{fig:betas_distribution}.}
All other experimental settings are kept the same as in Section~\ref{sec:medmnist_main}.
Table~\ref{tab:medmnist_beta} shows that all approaches are relatively robust to different $\beta$ values. 
This observation aligns with prior studies \citep{wang2023does} that highlight the robustness of SSL-based approaches to data heterogeneity in FL environments. 
This also indicates that LW-FedSSL and Prog-FedSSL exhibit robustness to data heterogeneity inherent in end-to-end SSL training.

\begin{table*}[ht]
\centering
\caption{Evaluation on the BloodMNIST dataset across different input dimensions and patch size settings.}
\label{tab:medmnist_input_size}
\resizebox{0.9\linewidth}{!}{
\begin{tabular}{@{}lrrrcccc@{}}
\toprule
\multicolumn{1}{c|}{}                   & \multicolumn{1}{c}{Memory (GB)} & \multicolumn{1}{c}{GFLOPs}               & \multicolumn{1}{c|}{Comm. (GB)} & Precision      & Recall         & F1             & Accuracy       \\ \midrule
\multicolumn{8}{c}{Image Size: $64 \times 64$, Patch Size: $4 \times 4$, Number of Patches: 256}                                                                                                                                                                       \\ \midrule
\multicolumn{1}{l|}{FedMoCoV3}          & 11.48 (1.00×)                                         & 2705 (1.00×)         & \multicolumn{1}{c|}{8.45 (1.00×)}                                         & 91.18          & 90.84          & 90.68          & 90.82          \\
\multicolumn{1}{l|}{LW-FedSSL (ours)}   & \textbf{3.10 (0.27×)}                                 & \textbf{640 (0.24×)} & \multicolumn{1}{c|}{\textbf{1.66 (0.20×)}}                                & 92.90          & 93.15          & 92.76          & 93.20          \\
\multicolumn{1}{l|}{Prog-FedSSL (ours)} & 11.46 (1.00×)                                         & 1468 (0.54×)         & \multicolumn{1}{c|}{5.09 (0.60×)}                                         & \textbf{94.12} & \textbf{93.73} & \textbf{93.77} & \textbf{93.62} \\ \midrule
\multicolumn{8}{c}{Image Size: $128 \times 128$, Patch Size: $8 \times 8$, Number of Patches: 256}                                                                                                                                                                     \\ \midrule
\multicolumn{1}{l|}{FedMoCoV3}          & 11.54 (1.00×)                                         & 2717 (1.00×)         & \multicolumn{1}{c|}{8.48 (1.00×)}                                         & 89.72          & 89.77          & 89.53          & 89.61          \\
\multicolumn{1}{l|}{LW-FedSSL (ours)}   & \textbf{3.15 (0.27×)}                                 & \textbf{645 (0.24×)} & \multicolumn{1}{c|}{\textbf{1.66 (0.20×)}}                                & 92.16          & 92.23          & 92.03          & 92.03          \\
\multicolumn{1}{l|}{Prog-FedSSL (ours)} & 11.52 (1.00×)                                         & 1479 (0.54×)         & \multicolumn{1}{c|}{5.13 (0.60×)}                                         & \textbf{94.41} & \textbf{93.65} & \textbf{93.87} & \textbf{93.61} \\ \midrule
\multicolumn{8}{c}{Image Size: $224 \times 224$, Patch Size: $16 \times 16$, Number of Patches: 196}                                                                                                                                                                   \\ \midrule
\multicolumn{1}{l|}{FedMoCoV3}          & 8.88 (1.00×)                                          & 2028 (1.00×)         & \multicolumn{1}{c|}{8.62 (1.00×)}                                         & 90.20          & 90.21          & 89.94          & 90.05          \\
\multicolumn{1}{l|}{LW-FedSSL (ours)}   & \textbf{2.84 (3.12×)}                                 & \textbf{487 (0.24×)} & \multicolumn{1}{c|}{\textbf{1.67 (0.19×)}}                                & 93.16          & 93.01          & 92.88          & 92.80          \\
\multicolumn{1}{l|}{Prog-FedSSL (ours)} & 8.87 (1.00×)                                          & 2028 (0.55×)         & \multicolumn{1}{c|}{5.26 (0.61×)}                                         & \textbf{95.23} & \textbf{95.26} & \textbf{95.13} & \textbf{95.14} \\ \bottomrule
\end{tabular}
}
\end{table*}

\begin{table}[ht]
\centering
\caption{Evaluation on the BloodMNIST dataset across different levels of data heterogeneity.}
\label{tab:medmnist_beta}
\resizebox{0.5\linewidth}{!}{
\begin{tabular}{@{}lcccc@{}}
\toprule
\multicolumn{1}{c|}{}                   & Precision      & Recall         & F1             & Accuracy       \\ \midrule
\multicolumn{5}{c}{$\beta=50$}                                                                              \\ \midrule
\multicolumn{1}{l|}{FedMoCoV3}          & 91.05          & 91.09          & 90.85          & 91.09          \\
\multicolumn{1}{l|}{LW-FedSSL (ours)}   & 92.60          & 92.92          & 92.56          & 92.77          \\
\multicolumn{1}{l|}{Prog-FedSSL (ours)} & \textbf{94.33} & \textbf{94.46} & \textbf{94.27} & \textbf{94.49} \\ \midrule
\multicolumn{5}{c}{$\beta=5$}                                                                               \\ \midrule
\multicolumn{1}{l|}{FedMoCoV3}          & 91.18          & 90.84          & 90.68          & 90.82          \\
\multicolumn{1}{l|}{LW-FedSSL (ours)}   & 92.90          & 93.15          & 92.76          & 93.20          \\
\multicolumn{1}{l|}{Prog-FedSSL (ours)} & \textbf{94.12} & \textbf{93.73} & \textbf{93.77} & \textbf{93.62} \\ \midrule
\multicolumn{5}{c}{$\beta=0.5$}                                                                             \\ \midrule
\multicolumn{1}{l|}{FedMoCoV3}          & 92.43          & 92.26          & 92.10          & 92.14          \\
\multicolumn{1}{l|}{LW-FedSSL (ours)}   & 92.55          & 93.15          & 92.68          & 92.93          \\
\multicolumn{1}{l|}{Prog-FedSSL (ours)} & \textbf{93.34} & \textbf{93.49} & \textbf{93.24} & \textbf{93.59} \\ \bottomrule
\end{tabular}
}
\end{table}

\section{Conclusion}
\label{sec:conclusion}

Edge devices in distributed environments often struggle to meet the computational and communication demands of federated self-supervised learning. 
In response to these challenges, we propose a layer-wise training approach named \mbox{LW-FedSSL}, which allows FL clients to perform self-supervised learning with significantly improved resource efficiency.
By dividing the training process into multiple stages and focusing on a subset of layers at each stage, LW-FedSSL effectively reduces resource consumption.
When applied to MoCoV3, LW-FedSSL requires $3.34 \times$ less memory, consumes $4.20 \times$  fewer GFLOPs, and reduces total transmission costs (download + upload) by $5.07 \times$, while maintaining comparable performance to its end-to-end counterpart, FedMoCoV3.
Furthermore, we explore a progressive training approach named \mbox{Prog-FedSSL}. 
Although Prog-FedSSL is less resource-efficient than LW-FedSSL, it remains more efficient than end-to-end training and achieves superior performance in most cases.
Through extensive experiments across various FL settings, datasets, and ablations, we demonstrate the effectiveness of LW-FedSSL and Prog-FedSSL. 
The results highlight the potential of layer-wise and progressive training strategies for enhancing the scalability and practicality of self-supervised learning in resource-constrained FL environments.

\appendix

\section{Convergence Behavior}
\label{sec:convergence}

\begin{figure*}[ht]
    \centering
    \subfloat[LW-FedSSL.]{\includegraphics[width=0.4\linewidth]{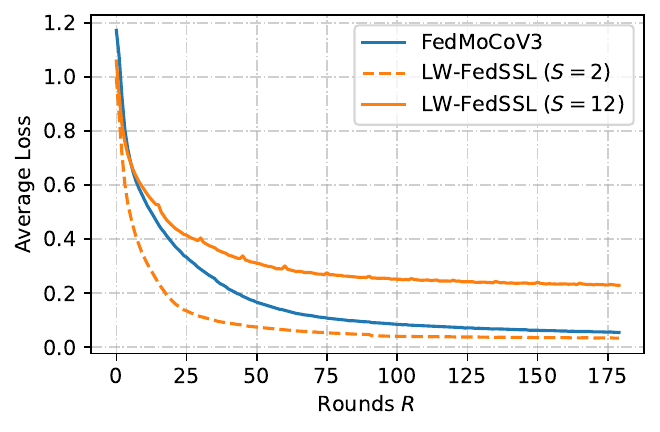} \label{subfig:loss_convergence_lw}} \hspace{2em}
    \subfloat[Prog-FedSSL.]{\includegraphics[width=0.4\linewidth]{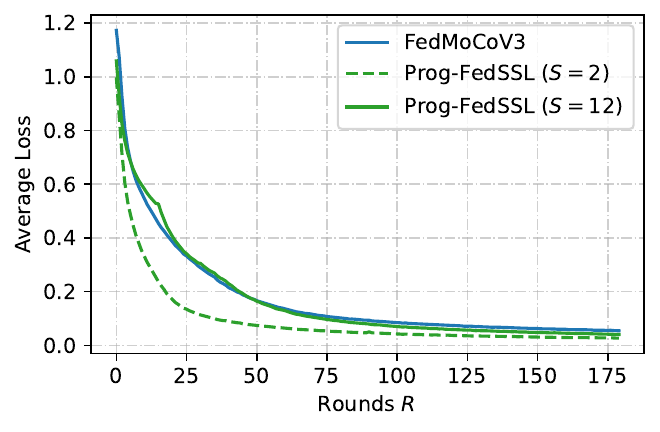} \label{subfig:loss_convergence_prog}}
    \caption{Average client loss over communication rounds for (a) LW-FedSSL and (b) Prog-FedSSL, using stage settings $S=2$ and $S=12$ (default), compared with end-to-end FedMoCoV3. LW-FedSSL converges in both settings, with $S=2$ exhibiting convergence behavior closer to FedMoCoV3. Prog-FedSSL converges similarly to FedMoCoV3 in both settings.}
	\label{fig:convergence}
\end{figure*}

To examine the convergence behavior of our proposed strategies, we plot the average client loss at each communication round for both LW-FedSSL and Prog-FedSSL.
Fig.~\ref{fig:convergence} shows the results for two settings: $S=2$ and $S=12$ (the default). 
For LW-FedSSL, we observe that both settings converge over the course of training. 
However, with $S=12$, the loss remains slightly higher than that of FedMoCoV3, while reducing the number of stages to $S=2$ results in convergence behavior closer to FedMoCoV3.
This is expected since FedMoCoV3 can be regarded as training with one stage (i.e., $S=1$).
For Prog-FedSSL, both stage settings demonstrate convergence patterns similar to FedMoCoV3.

\section{Additional Details}
\label{sec:additional_details}

\subsection{Details on FLOPs Calculation}
\label{sec:calculate_flops}

Here, we discuss the details on FLOPs calculation used in our experiments.
In layer-wise training, calculating FLOPs for inactive (frozen) layers involves simply considering FLOPs associated with the forward pass. 
However, for active layers, the calculation must account for both the forward and backward passes. 
While numerous works have addressed FLOPs calculation for the forward pass (inference), limited studies are available on the practice of FLOPs calculation for the backward pass.
Notably, existing related works \citep{wang2022progfed,pengfei2023towards} also do not provide specific details on how FLOPs are computed.
Some studies \citep{estimate,whattheback,aiandcompute,epoch2021backwardforwardFLOPratio} suggest that the number of operations in a backward pass of a neural network is often twice that of a forward pass.
Following these studies, we adopt a backward-forward ratio of 2:1 to calculate the total FLOPs for active layers.
We use the \texttt{FlopCountAnalysis} tool in the fvcore \cite{fvcore} library for calculating the FLOPs of the forward pass. 
We only consider a single input sample for FLOPs calculation.

\begin{figure*}[ht]
    \centering
    \subfloat[$\beta=50$]{\includegraphics[width=0.235\linewidth]{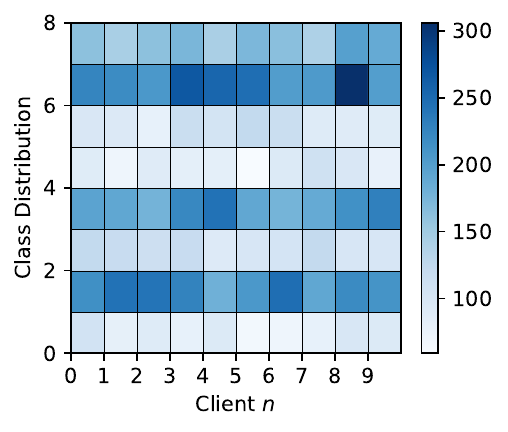} \label{subfig:beta_50}} \hspace{1em}
    \subfloat[$\beta=5$]{\includegraphics[width=0.235\linewidth]{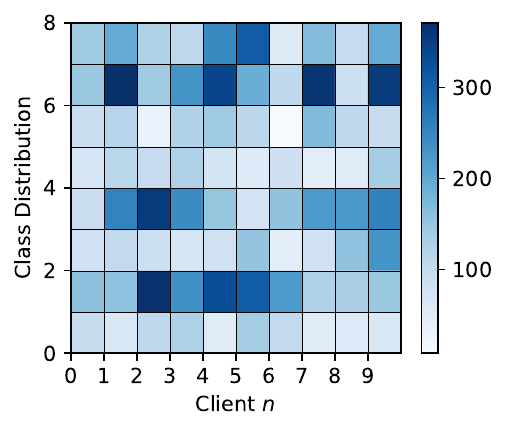} \label{subfig:beta_5}} \hspace{1em}
    \subfloat[$\beta=0.5$]{\includegraphics[width=0.235\linewidth]{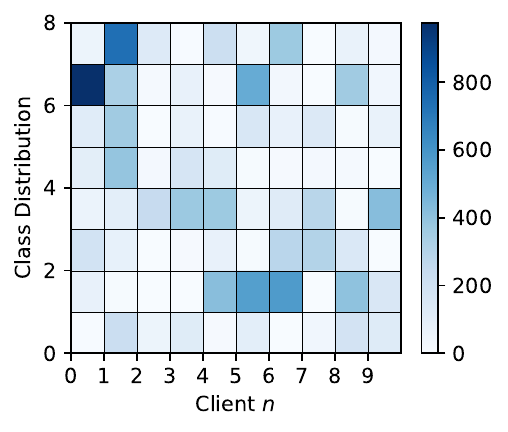} \label{subfig:beta_0.5}} 
	\caption{Client data distribution with different $\beta$ values for the BloodMNIST dataset. A darker color denotes a higher number of data samples for a specific class in a client.}
	\label{fig:betas_distribution}
\end{figure*}

\begin{figure*}[ht]
    \centering
    \includegraphics[width=.98\linewidth]{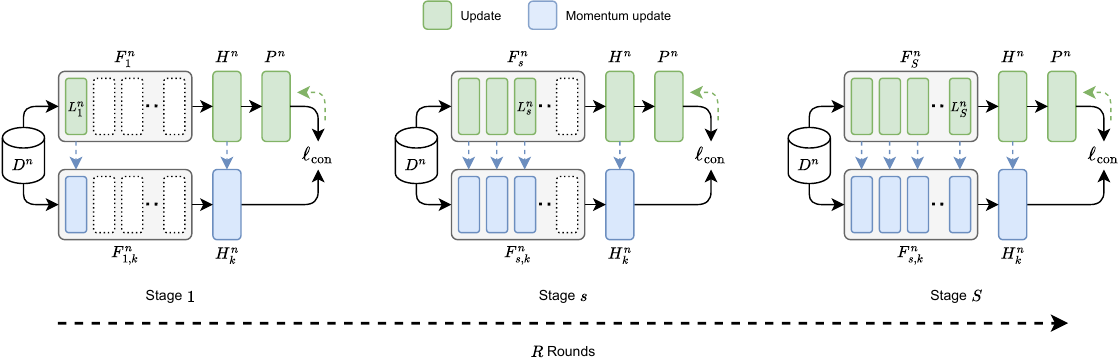}
    \caption{Prog-FedSSL: Local training process across different stages for the $n$-th client. At the beginning of each stage $s \in [1,S]$, a new layer $L_s$ is sequentially added to the encoder $F_{(s-1)}$, increasing its depth. During stage $s$, Prog-FedSSL updates all existing layers (i.e., $[L_1, \dots, L_s]$) within $F_s$.}
    \label{fig:prog_fedssl}
\end{figure*}

\subsection{Data Distribution}

Fig.~\ref{fig:betas_distribution} illustrates the data distribution among clients for different $\beta$ values used in Section~\ref{sec:data_heterogeneity}. Lower $\beta$ values correspond to increasingly heterogeneous data distributions among clients.



    



            



            



\begin{algorithm}[ht]
\caption{Prog-FedSSL (\emph{Server-side})}
\label{alg:progfedssl_server}
\textbf{Input:} encoder $F_0$, projection head $H$, prediction head $P$, number of stages $S$, number of rounds per stage $R_s$ where $s \in [1,S]$ \\
\textbf{Output:} encoder $F_S$

\begin{algorithmic}[1]
    \State \textbf{Server executes:}
    \State Distribute $F_0$ to clients
    \For{stage $s=1,2,\dots, S$}
        \State Initialize new layer: $L_s$
        \State $F_s \leftarrow$ Attach $L_s$ to $F_{(s-1)}$
        \For{round $r=1,2,\dots, R_s$}
            \For{client $n=1,2,\dots,N$ in parallel}
                \vspace{0.2em}
                \State $F_s^n, H^n, P^n \leftarrow \text{Train}(n, F_s, H, P)$
                \vspace{0.2em}
                \State $w^n = \frac{|D^n|}{|D|}$, where $D=\bigcup_{n=1}^{N} D^n$
            \EndFor
            \vspace{0.2em}
            \State $F_s \leftarrow \sum_{n=1}^N w^n F_s^n$
            \vspace{0.2em}
            \State $H \leftarrow \sum_{n=1}^N w^n H^n$
            \vspace{0.2em}
            \State $P \leftarrow \sum_{n=1}^N w^n P^n$
        \EndFor
    \EndFor
    \State return $F_S$
\end{algorithmic}
\end{algorithm}

\begin{algorithm}[H]
\caption{Prog-FedSSL (\emph{Client-side})}
\label{alg:progfedssl_client}
\textbf{Input:} local dataset $D^n$, number of local epochs $E$, momentum $\mu$, temperature $\tau$\\
\textbf{Output:} encoder $F^n_s$, projection head $H^n$, prediction head $P^n$ 
\begin{algorithmic}[1]
    \State \textbf{Client executes:} Train$(n, F_s, H, P)$:
        \vspace{0.2em}
        \State Initialize: $F^n_s \leftarrow F_s$, $H^n \leftarrow H$, $P^n \leftarrow P$
        \vspace{0.2em}
        \State Target branch: $F^n_{s,k} \leftarrow F^n_s$, $H^n_k \leftarrow H^n$ 
        \vspace{0.2em}
        \For{epoch $e=1,2,\dots,E$}
            \For{each batch $x \in D^n$}
                \State $x_1 \leftarrow \text{{Augment}}(x)$
                \vspace{0.2em}
                \State $x_2 \leftarrow \text{{Augment}}(x)$
                \vspace{0.2em}
                
                \State $q_1 \leftarrow P^n(H^n(F^n_s(x_1)))$
                \vspace{0.2em}
                \State $q_2 \leftarrow P^n(H^n(F^n_s(x_2)))$
                \vspace{0.2em}

                \State $k_1 \leftarrow H^n_k(F^n_{s,k}(x_1))$
                \vspace{0.2em}
                \State $k_2 \leftarrow H^n_k(F^n_{s,k}(x_2))$
                \vspace{0.2em}
                
                \State $\mathcal{L} \leftarrow \ell_\text{con}(q_1, k_2, \tau) + \ell_\text{con}(q_2, k_1, \tau)$
                \Comment{Local loss}
                \vspace{0.2em}

                \State $F^n_s, H^n, P^n \leftarrow$ Update with $\nabla\mathcal{L}$
                \vspace{0.2em}

                \State $F^n_{s,k}, H^n_k,\leftarrow$ Momentum update with $\mu, F^n_s, H^n$
                \vspace{0.2em}
            
            \EndFor
        \EndFor
        \State return $F^n_s, H^n, P^n$
\end{algorithmic}
\end{algorithm}





\bibliographystyle{elsarticle-num} 
\bibliography{main}




\end{document}